\documentclass[journal]{IEEEtran}
\ifCLASSINFOpdf
\else
\fi
\usepackage{comment}
\usepackage{hyperref} 
\usepackage{graphics}
\usepackage{algorithm}
\usepackage{algpseudocode}
\usepackage{subfigure}
\usepackage{xcolor}
\usepackage{nomencl}
\makenomenclature

\usepackage{amssymb}
\usepackage{pifont}

\usepackage[cmex10]{amsmath}
\usepackage{mathabx}
\usepackage{array}
\usepackage{mdwmath}
\usepackage{mdwtab}
\usepackage{url}
\usepackage{comment}
\usepackage{float}
\usepackage{listings}
\usepackage[compact]{titlesec}
\restylefloat{table}
\usepackage{amsmath, amsthm, amssymb}
\usepackage[ansinew]{inputenc}
\usepackage{mathtools}
\begin{document}
%
\title{
Bayesian Reinforcement Learning for Automatic Voltage Control under Cyber-Induced Uncertainty
}
%
%
%

\author{Abhijeet Sahu,~\IEEEmembership{Student Member,~IEEE,}
        and~Katherine~Davis,~\IEEEmembership{Senior Member,~IEEE}
}

%
%

\markboth{IEEE December~2022}%
{Shell \MakeLowercase{\textit{et al.}}: Bare Demo of IEEEtran.cls for IEEE Journals}
%



\maketitle

\begin{abstract}
Voltage control is crucial to large-scale power system reliable operation, as timely
reactive power support
can help prevent
widespread outages. However, there is currently no built in mechanism for power systems to ensure that the voltage control objective to maintain reliable operation will survive or
sustain the uncertainty caused under adversary presence.
%
Hence, this work introduces a Bayesian Reinforcement Learning (BRL) approach for power system control problems, with focus on sustained
voltage control under uncertainty in a cyber-adversarial environment.
%
%
This work proposes a data-driven BRL-based approach for automatic voltage control by formulating and solving a Partially-Observable Markov Decision Problem (POMDP), where the states are 
partially observable due to cyber intrusions. 
The techniques are evaluated 
on the WSCC and IEEE 14 bus systems.  Additionally, BRL techniques assist in automatically finding a threshold for exploration and exploitation in various RL techniques. 
\end{abstract}


%
\IEEEpeerreviewmaketitle
\nomenclature{$Q(s,a)$}{State Action Value Function}
\nomenclature{$s_t$}{Current State}%
\nomenclature{$s_{t+1}$}{Next State}%
\nomenclature{$V(s)$}{State Value Function}%
\nomenclature{$\gamma$}{Discount Factor}
\nomenclature{$b_{t}(s)$}{Belief State at time t}
\nomenclature{$\Omega$}{Observation Probability Distribution}
\nomenclature{$P(s'|s)$}{Transition Probability}
\nomenclature{$R(s,a)$}{Reward function}
\nomenclature{$\phi_{s,s'}^{a}$}{Hyper state}
\nomenclature{$\mathcal{A}$}{Action Space}
\nomenclature{$\mathcal{S}$}{State Space}
\nomenclature{$\mathcal{O}$}{Observation Space}
\nomenclature{$\theta$}{Policy parameters}
\nomenclature{$k(x,x')$}{Kernel Functions}
\nomenclature{$\mathcal{D}_M$}{Datasets with M rows}
\nomenclature{$\xi$}{Trajectory}
\nomenclature{$KL(p,q)$}{Kullback-Liebler Divergence of distribution $p$ \& $q$}
\nomenclature{$VPI(s,q)$}{Value of Perfect Information}
\nomenclature{$G(\theta)$}{Fisher Information Matrix}
\nomenclature{$k_F$}{Invariant Fisher Kernel}
\nomenclature{$u(z;\theta)$}{Score Function}
\nomenclature{$\beta$}{Policy Learning Rate}
\nomenclature{$\nu$}{Parameterized Policy}
\nomenclature{$z$}{State-Action pair}
\nomenclature{$P_{t}^{\nu}(z_t)$}{$t$-step State Action occupancy density of policy $\nu$}
\nomenclature{$N$}{Number of voltage discretization level}
\nomenclature{$n_b$}{Number of Buses monitored}
\nomenclature{$M$}{Number of generators}
\nomenclature{$p$}{Number of action levels}
\nomenclature{$M_p$}{Possible actions}
\nomenclature{$n_v$}{Number of buses with voltage violation}
\nomenclature{$t_p$}{True observation probability}
\nomenclature{$r_p$}{Residual observation probability}
\nomenclature{$\hat{S}_t(z,z')$}{Covariance of Posterior distribution $Q(z)$}
\nomenclature{$\hat{Q}_t(z,z')$}{Mean of Posterior distribution $Q(z)$}
\nomenclature{$\nabla_{\boldsymbol{\theta}} \eta(\boldsymbol{\theta})$}{Gradient of expected return}
\printnomenclature

\section{Introduction}
%
%
%
%

 

The increasing complexity of large interconnected power systems makes it difficult to ensure normal grid operation. While automatic controls
are deployed to help operators maintain the power grid's reliability and security, a major challenge is 
how to handle operational
uncertainty caused due to cyber intrusions.
The cyber defense problem in power systems can hence be framed as a problem of decision making under uncertainty.  Reinforcement Learning (RL) 
with Bayesian inference can help
sustain operational reliability under threat. 

In an RL problem, an agent interacts with an environment 
to find an action-selection strategy 
to optimize its long-term reward. Bayesian Reinforcement Learning (BRL) is a type of RL that
uses Bayesian inference to incorporate information into the learning process. It incorporates
prior probabilistic distributions and new information,
using standard rules of Bayesian inference,
and may be model-based or model-free. 

The grid's cyber infrastructure enables timely real-time monitoring and actuation of controls. The concern is that 
challenges such as network congestion, cyber intrusions, and increased misconfigurations may prevent timely and accurate
operations. 
To address this, adversarial RL techniques have been proposed
~\cite{cyber_rl_2,cyber_rl_1,cyber_rl_3}, with a cyber intruder modeled as an adversarial agent and a defender agent, as in \textit{CyberBattleSim}~\cite{microsoft_cbs}. 
The main contribution of this paper 
is to 
present a novel approach to solve the control problem under cyber-induced uncertainty, by detailing a complete framework for 
modeling a Partially Observable Markov Decision Process (POMDP) based environment for decision-making problems, as the state is not completely observable.


BRL has proven effectual in solving other POMDP problems, e.g.,~\cite{brl_pomdp,brl_pomdp2}.  A Factored Bayes-Adaptive POMDP 
is proposed in~\cite{brl_pomdp} that learns the dynamics of partially observable systems through the
Monte-Carlo Tree Search algorithm, 
while ~\cite{brl_pomdp2} proposes 
BRL for solving POMDPs with a continuous state space. 
%
%
BRL intrinsically solves the exploration-exploitation problem, as well as incorporates regularization, by avoiding the problem of a few outliers steering it away from true parameters. A frequentist approach of handling parameter uncertainty is computationally infeasible and also requires more data, since to measure uncertainty, it relies on a null hypothesis and confidence intervals (CI), where $CI = x'\pm z \frac{s}{\sqrt{n}}$ and $x'$ is the sample mean, $z$ is the confidence level value, $s$ is the sample standard deviation and $n$ the sample size.
BRL has been utilized in 
cyber-security management~\cite{brl_cybersec}, as well as MDPs with model uncertainty~\cite{brl_uncertainty}. In~\cite{brl_cybersec}, BRL is introduced for faster intrusion detection, leveraging limited scan results and warnings.
For robust adaptation 
to the real world dynamic environment,
a continuous Bayes-Adaptive MDP is proposed~\cite{brl_uncertainty} and solved using Bayesian Policy Optimization. Unlike the prior works on cyber-security management, this work addresses
the voltage control problem in power systems that can encounter cyber-induced data manipulation in the OT network. 

The major contributions of this paper are:

\begin{itemize}
    
    \item Creation of a cyber-physical POMDP problem for automatic voltage control under uncertainty caused due to FDI attack.
    \item  Propose the use of the BRL techniques to solve the POMDP and Bayesian variants of Deep Q Network (DQN) Learning to improve the performance in comparison to conventional DQN technique.
     \item Evaluation of the impact of varying prior distribution of the Q function and other parameters of BRL techniques considered for the WSCC and IEEE 14 bus case.
\end{itemize}

\section{Background}

BRL facilitates the use of explicit probability distributions, which allows uncertainty quantification and enables improved learning strategies. BRL implicitly finds the exploitation/exploration tradeoff and reduces risks in decision making. 
Exploitation is a greedy approach, where agents try to make the best decision to get more rewards by using an estimated value from the current information. In exploration, agents focus on improving their knowledge about each action
to obtain long-term benefits.

In the Bayesian approach, the notion of how certain we are about a state action value function, $Q(s,a)$, is represented through the distribution of $P_{Q}(s,a)$, which allows us to quantify how much we know about the $s,a$ combination and its reward.
A high variance 
makes the agent explore more, while a low variance provides more certainty; hence, such a combination may not be worth exploring, if a more optimal action is already deciphered by the agent.
Moreover, due to its probabilistic framework, one can encode domain knowledge through the prior distribution in BRL. BRL also facilitates regularization by preventing exceptions in the dataset from causing over-fitting problems while learning. Amidst its multiple advantages, the major concern of BRL is that it is computationally  expensive, since here we make an estimation of the distribution parameters, unlike point estimates in other RL methods. Utilizing faster algorithms such as~\cite{faster_algo} that performs High-Dimensional Robust Covariance Estimation can improve the performance but currently is out of scope for discussion in this paper.

Bayesian Stackelberg Markov Games (BSMG) have been leveraged in a proactive Moving Target based Defense (MTD) mechanism for getting rid of an adversary's reconnaissance stage~\cite{mtd_bayesian}. Given the adversary competence is unknown, it is inadequate to model the leader-follower games conventionally modeled in MTD problems. Hence, to identify optimal sequential strategies, BSMG are formulated and solved using Bayesian Strong Stackelberg Q-Learning and validated for web-application security in ~\cite{mtd_bayesian}. 

The automatic voltage control problem has been studied rigorously 
(e.g., review paper~\cite{low_review}), that use various value and policy learning techniques from RL, but the voltage regulation problems are formulated as an MDP. \textit{\textbf{This work is the first attempt to formulate
the voltage control problem as a POMDP and solve using BRL.}}



\section{Problem Formulation}\label{sec:formulation}

\subsection{Automatic Voltage Control}
Voltage control in power systems is primarily performed through excitation control or voltage regulators at generating stations, through tap changing transformers, induction regulators, shunt reactors and capacitors or synchronous condensers. The automatic voltage control considered in this work is related to the excitation control in which when the terminal voltage drops, the field current is increased to compensate for the voltage drop. A automatic voltage regulator (AVR) detects the terminal voltage and compares it with the reference voltage, with the goal to control the excitation voltage of the generator to cancel out the error (i.e. difference between reference and terminal voltage).

\begin{figure}[h]
\centerline{\includegraphics[scale=0.9]{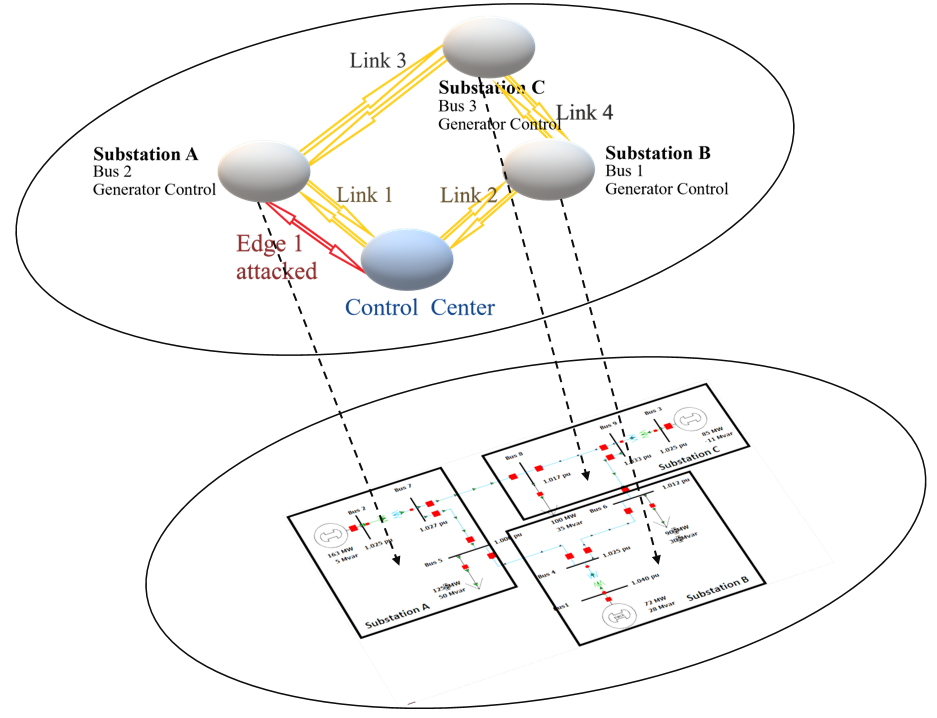}}
\caption{WSCC 9-Bus Cyber Physical Model}
\label{fig:phy_sys}
\vspace{-.1in}
\end{figure}

A voltage control problem is formulated, 
for a three-substation network based on the WSCC 9-bus case~\cite{wecc_9}, with 4 broadcast domains, one for each substation and one for the control center, shown in Fig.~\ref{fig:phy_sys}. The voltages of the 3 generators are regulated, during cyber intrusions targeting FDI attack on the phasor measurements. The control actions are defined as the reference voltage setpoint given to the AVR. 

\vspace{-1mm}



\subsection{Cyber Threat Model}
Numerous works on False Data Injection (FDI) attacks and defense exist
, such as~\cite{KUNDU_a3d} and~\cite{gnn_osman},  on
power system state estimation.
For FDI attacks,
an adversary is considered to be
modifying measurements of 
devices such as in supervisory control and data acquisition (SCADA) 
or phasor measurement units (PMUs),  
to change 
readings to target the 
miscontrol of 
power,
based on specific targets.
A previous work incorporated FDI in an emulated enviornment with DNP3 and defense using machine learning%
~\cite{reslab}%
~\cite{data_fusion}. Unlike the previous works on detection, this work aims to control the voltage under cyber threats, through a data-driven 
RL approach.
Hence, this work follows a unique technique of incorporating the FDI through defining an \textit{observation model} in the POMDP space, where the control RL agent observes a different value from the actual state, following a probability distribution.
The observation model in the POMDP, $\Omega(o|s)$, is introduced in Section~\ref{pomdp_avr}.
\subsection{MDP model}

The proposed 
automatic voltage control under cyber adversarial presence is based on a Markov Decision Process (MDP).
A MDP is a discrete-time stochastic process used to describe the agent and environment interactions. 
In our problem, the agents are compared to the decision making engines, and the environment is the current state of the cyber-physical system, defined by a tuple of five components: states ($S$), actions ($A$), state transition model $P(s_{t+1}|s_{t},a_{t})$ 
for action $a$ in state $s$, reward model $R(s_{t+1}|s_{t},a_{t})$
an agent receives 
after execution,
and discount factor $\gamma$ that controls future rewards.  

The value function $V(s)$ is the benefit for the agent to be in state $s$,
or the expected total reward for an agent starting from state $s$; it depends on the policy $\pi$ used to
pick actions.
The state-action pair function called the Q-function is also considered. The optimal Q-function $Q^{*}(s, a)$ 
is an indication of how good it is for an agent to pick action $a$ while in state $s$, 
equal to the summation of immediate reward after performing action $a$ while in state $s$ and the discounted expected future reward after the transition to a next state $s'$:
\begin{equation}\label{eq:3}
    \begin{aligned}
   Q^{*}(s, a)=R(s, a)+\gamma \sum_{s^{\prime} \in \mathbb{S}} p\left(s^{\prime} | s, a\right) V^{*}\left(s^{\prime}\right)
    \end{aligned}
\end{equation}



Next, we describe modifications to this model to account for cyber induced uncertainty by modeling POMDPs. The POMDP can be modeled using notions of both \textit{belief states} and \textit{hyper states}~\cite{brl_survey}. In the upcoming section, both variants are introduced.

\subsection{Bayes-Adaptive MDP model}
The BAMDP~\cite{bamdp} is 
an extension of the 
MDP and also considered to be a special case of POMDP~\cite{brl_survey}.
The state space of the BAMDP combines jointly the initial
set of states S, with the posterior parameters on the transition
function $\Phi$, defined collectively as the \textit{hyper-state}.

The Bellman equation for the BAMDP is 
the following:
\begin{equation}
\label{eq:bamdp}
    \begin{aligned}
V^{*}(s, \phi) 
&=\max _{a \in \mathcal{A}}\left[R(s, a)+\gamma \sum_{s^{\prime} \in \mathcal{S}} \frac{\phi_{s, s^{\prime}}^{a}}{\sum_{s^{\prime \prime} \in \mathcal{S}} \phi_{s, s^{\prime \prime}}^{a}} V^{*}\left(s^{\prime}, \phi^{\prime}\right)\right]
\end{aligned}
\end{equation}
 The BAMDP is solved using Bayesian Q-Learning (BQL) (Section~\ref{bql}). For the AVC problem, we have discretized the voltage levels from 0.9 to 1.1 p.u. into $N$ levels, because the number of $Q(s,a)$ distributions to be learned needs to be finite.
 %
 The action space is also discretized for $M$ generators in the system, with $p$ levels in the range of 0.95-1.05 p.u., resulting in $M^{p}$ possible actions. Hence the number of $Q$ distributions to learn is $M^{p} \times N^{n_b}$, where $n_b$ is the number of buses monitored. The reward model for the BAMDP is as follows,
\begin{equation}\label{step_reward}
    R = 50 - 100*n_v
\end{equation}
where $n_v$ is the number of buses with violation i.e ($V \ge 1.05$ or $V \le 0.95$). 
Assuming $|S|$ states at the start of an episode and a fully-connected state space model, the BAMDP will have $|S|^t$ reachable states at horizon $t$. Hence, based on $t$, there is a computational challenge in representing states using \textit{hyper states}~\cite{brl_survey}. Hence, the belief-state variant of POMDP is considered.

\subsection{POMDP model}
In a POMDP~\cite{pomdp}, the system state is not observable. A POMDP is defined through a tuple $<S,A,O,P,\Omega,P_{0},R>$, where $S$ is the set of states, $A$ is the set of actions, $O$ is the set of observations, $P(.|s)$ gives the probability distributions over next states, for action $a$ at state $s$, $\Omega$ gives the probability distributions over possible observations, for $a$ at $s$, $P_{0}$ is the probability distribution of initial states, and $R(s,a)$ is a random variable 
for reward obtained by taking $a$ from $s$. 
Since the states are not available, 
the state space is inferred from the observation and action sequences. This is defined as
the information or belief state,
updated as follows:
\begin{equation}\label{pomdp_eq1}
\begin{aligned}
b_{t+1}\left(s^{\prime}\right)=\frac{\Omega\left(o_{t+1} \mid s^{\prime}\right) \int_{\mathcal{S}} P\left(s^{\prime} \mid s\right) b_{t}(s) d s}{\int_{\mathcal{S}} \Omega\left(o_{t+1} \mid s^{\prime \prime}\right) \int_{\mathcal{S}} P\left(s^{\prime \prime} \mid s\right) b_{t}(s) d s d s^{\prime \prime}}
\end{aligned}
\end{equation}

Solving a POMDP obtains
an optimal policy $\nu_{*}$ that maximizes the discounted sum of rewards over all information states,
through the Bellman equation, where $V^*(b)$ equals:

\begin{equation}\label{pomdp_eq2}
\begin{aligned}
\max _{a \in \mathcal{A}}\left[\int_{\mathcal{S}} R(s, a) b(s) d s+\gamma \int_{\mathcal{O}} \operatorname{Pr}\left(o \mid b, a\right) V^{*}\left(\tau\left(b, a, o\right)\right) d o\right]
\end{aligned}
\end{equation}

and $\tau(b_{t}, a, o)$ is the $b_{t+1}$ in Eq.~\ref{pomdp_eq1}.

\subsection{POMDP-based AVC Problem}\label{pomdp_avr}
The belief states are updated based on 
Eq.~\ref{pomdp_eq1}.
The observation model 
for the voltage regulation problem depends on 
the true observation probability $t_p$ and the residual probability $r_{p}$. For example,
if a state is 7 (i.e., the voltage value is between 0.96 - 0.97 p.u., for $N = 20$), the probability of observing state 7 will be $t_p$, while the probability of observing its neighbor states
6 or 8 are $(1 - t_{p} - r_{p})/2$ each, and the probability of observing the rest of the possible states is distributed 
with $r_p$. For voltage levels in 
0.95 - 1.05 p.u., $r_{p}$ is kept higher than in the rest of the voltage regions ($r_p(s') \ge r_p(s'')$,  where $s' \in (0.95,1.05)$ and $s'' \ge 1.05 \cup s'' \le 0.95 $). Such formulation
keeps the variance of the distribution high for major operating regions.
%
\begin{equation*}
\Omega(o|s) = \begin{cases}
                t_p \quad &\text{if} \, o = s\\
                1 - t_p - r_p(s) \quad &\text{if} \, o \in {s-1,s+1}\\
                r_p(s) \quad & elsewhere
            \end{cases}
\end{equation*}

Based on number of voltages monitored $n_b$, the observation and state space increase exponentially as $N^{n_b}$. In the results, we analyze the computation complexity of the BRL agents when the state-space increases. The prior of the transition model follows a Dirichlet distribution, as described~\cite{BDP}.



Then, based on the discussions and notations above,
the reward model for the POMDP is the following,
\begin{equation}
    R(s) = 1- P(.|s) + P(.|s)*R_{orig}(s)
\end{equation}

\noindent where $R_{orig}$ is the actual reward obtained based on Eq.~\ref{step_reward}.

\section{Solution of Proposed Formulation}

For solving the models in
Sec.~\ref{sec:formulation}, techniques from
Bayesian Reinforcement Learning are considered. 
An RL algorithm may be model-based or model-free; it may be based on 
value function, policy search, or a mixed-model approach.
Model-based RL explicitly learns the system model (transition probabilities, reward model) and solves the MDP, while model-free RL uses sample trajectories through interaction with the environment. Some RL algorithms learn the value function, such as value iteration, Q-learning, and SARSA~\cite{rl_book}. Policy search solves the MDP by direct exploration of the policy, which can grow exponentially with state space size. Hence, policy gradient methods help by
defining a differentiable function of a parameterized vector, and the policy is searched as directed by the gradient.
%
Mixed-model 
techniques include
Actor-Critic~\cite{actor_critic} that uses policy gradient followed by evaluating the policy, estimating a value function.

In BRL,
model-based techniques include Bayesian Dynamic Programming, Value of Perfect Information (VPI), and tree-search Q function approximation methods (Bayesian Sparse Sampling, Branch-and Bound Search). Model-free includes Bayesian Deep Q-Network Learning (BDQN), Gaussian Process Temporal Differencing (GPTD), Bayesian Policy Gradient based, and Bayesian Actor-Critic (BAC)
~(Fig 1.1 of \cite{brl_survey}). To solve the proposed problem, we adopt
VPI-based Q-learning (model-based), and
BDQN and BAC 
(model-free).


\subsection{Bayesian Q-Learning~\cite{bayesian_qlearn} 
} \label{bql}
In Bayesian Q-Learning, instead of solving for the value function using Eq.~\ref{eq:bamdp} for BAMDP,
a Dirichlet distribution is used to obtain the state-action values $Q^{*}(s,a)$ to select the action with highest expected return and value of perfect information (VPI)%
~\cite{bayesian_qlearn} ($E[q_{s,a}] + VPI(s,a)$). This estimates
expected improvement in policy by myopically looking over a single step horizon, following an exploration action. 
VPI is given as follows: 

\begin{equation}\label{eqn:vpi}
\operatorname{VPI}(s, a)=\int_{-\infty}^{\infty} \operatorname{Gain}_{s, a}(x) \operatorname{Pr}\left(q_{s, a}=x\right) d x
\end{equation}

\begin{equation}
\operatorname{Gain}\left(Q_{s, a}^{*}\right)= \begin{cases}\mathrm{E}\left[q_{s, a_{2}}\right]-q_{s, a}^{*} & \text { if } a=a_{1} \; q_{s, a}^{*}<\mathrm{E}\left[q_{s, a_{2}}\right] \\ q_{s, a}^{*}-E\left[q_{s, a_{1}}\right] & \text { if } a \neq a_{1} \; q_{s, a}^{*}>\mathrm{E}\left[q_{s, a_{2}}\right] \\ 0 & \text { otherwise }\end{cases}
\end{equation} 

The number of posterior distributions learned
is $2 \times N_{a} \times N_{s}$, where 2 is due to estimating mean and variance, and $N_a$ and $N_s$ are the numbers of actions and states, respectively. For the prior, we initialize a random distribution for the mean and variance of the Q. 
The \textit{PyMC3} library is used for training the model. After the prior is taken in the model, the likelihood is computed based on the observation of Q values  
from the environment. A Gaussian likelihood is considered.
\textit{Variational Inference}~\cite{vi} with the Adam optimizer is used for obtaining the posterior. The automatic differentiation variational inference works on the principal of finding the optimal posterior distribution $q^{*}$, given by the following,
\begin{equation}\label{min_kld}
q^{*}=\operatorname{argmin}_{q \in Q} f(q(\cdot), p(\cdot \mid y))
\end{equation}
where $f$ is the distance between 
$q$ and $p$ distributions. In variational inference theory, $f$ is the Kullback-Liebler Divergence (KL-Div), where $y$ is the observations/data, and $\theta$ is the parameter to be estimated:
\begin{equation}\label{kld}
\begin{aligned}
\mathrm{KL}(q(\cdot) \| p(\cdot \mid y)) \longrightarrow \int q(\theta) \log \frac{q(\theta)}{p(\theta \mid y)} d \theta \\
=\int q(\theta) \log \frac{q(\theta) p(y)}{p(\theta, y)} d \theta=\log p(y)-\int q(\theta) \log \frac{p(\theta, y)}{q(\theta)} d \theta
\end{aligned}
\end{equation}

$KL \ge 0$, and the second expression of the lower-right term 
in Eq.~\ref{kld} is 
the evidence lower bound (ELBO). 
Eq.~\ref{min_kld} is equivalent to maximizing the ELBO. 
Variational inference works well for Gaussian distributions.
For estimating multi-modal distributions, one must consider more advanced techniques such as Stein-Variation Gradient Descent~\cite{svgd} which generates an approximation based on a large number of particles.

\subsection{Bayesian Deep Q Network Learning (BDQN)}\label{bdqn_learn}

The pseudo code for both DQN and BDQN 
is given in Alg.~\ref{bdqn}.
%
BDQN combines both MCMC Sampling with the DQN, where the posterior sampling is carried out over the model parameters of the Q network.  This is the implementation of the proposed BDQN framework from~\cite{neal_thesis}. In a conventional neural network, weights and biases are found that minimize the error function using gradient-descent optimization called backpropagation with some penalty terms to the loss function for regularization i.e. to prevent overfitting. In the Bayesian approach, instead of training the weights by backpropagation, the weights are proposed based on sampling from Gaussian distribution a proposed weight and evaluating the likelihood $LL(w_p)$ and prior($PL(w_p)$) based on the output of Q network with proposed weight $w_p$; further utilizing the Metropolis-Hastings algorithm of computing the acceptance ratio ($r$), based on which performing the soft update on the target network parameter as performed in DQN. 
The acceptance ratio, $r$, is the ratio of the likelihood and prior probability as shown in Line 24 of the Alg.~\ref{bdqn}, the pseudo code for both DQN and BDQN. In the algorithm, the Line 20-27 is pertaining to the BDQN, while in DQN, Line 16-17 are executed, and the rest of the code is common to both the algorithms.

\begin{algorithm}[h]
	\caption{Pseudo Code for BDQN}
	\begin{algorithmic}[1]
		\Function{$BDQN$}{$D, DQN, BDQN$}
		\State Initialize $Q_{\theta}$ and Target $Q_{\theta^{'}}$ Network
		\For {each iteration}
		      \For {each environment step }
		            \State Act $a_t$ through $\epsilon$-greedy
		            \State Get $r_t,s_{t+1}$
		            \State Store $(s_t,a_t,r_t,s_{t+1})$ in $D$
	            \EndFor
	           \For {each update step} 
	                \State Initialize $w$, weight of the $Q$ network
	                \State Compute $LL(w)$ and $PL(w)$
	                \State sample $e_t = (s_t,a_t.r_t,s_{t+1}) from D$
	                \State Compute target Q value
	                \State $Q^{*}(s_t,a_t) = r_t + \gamma*Q_{\theta}(s_{t+1},argmax_{a'}Q_{\theta'}(s_{t+1},a'))$
	                \If {DQN}
	                    \State Gradient descent $(Q^{*}(s_t,a_t) - Q_{\theta}(s_t,a_t))^{2}$
	                    \State Update target N/W $\theta' = \tau*\theta + (1-\tau)*\theta'$
                    \EndIf
                    \If {BDQN}
	                \For {given sample length}
	                    \State $w_p$ sample a weight from $N(0,\sigma)$
	                    \State $\tau_p$ sample a $\tau$ from $N(0,\sigma)$
	                    \State Compute $LL(w_p)$ and $PL(w_p)$
	                    \State Acceptance Ratio $r= min(0,(LL(w_p)-LL(w)) + (PL(w_p)-PL(w)))$
	                    \If {$r \ge U(0,1)$}
	                        Update target network: $\theta' = \tau_p*\theta + (1-\tau_p)*\theta'$
	                    \EndIf
	                \EndFor
	                \EndIf
              \EndFor
          \EndFor
		\EndFunction
	\end{algorithmic}
	\label{bdqn}
\end{algorithm}

\subsection{Bayesian Policy Gradient \& Actor-Critic Agent}

In RL, the policy gradient methods design a parameterized policy and update the parameter through the performance gradient estimate. REINFORCE~\cite{rl_book} was the first Monte-Carlo based policy gradient technique used in the conventional RL approach, but due to its high variance, other techniques such as Actor-Critic, Advantageous Actor Critic, Proximal Policy Optimization, Natural Policy Gradient, etc., techniques are adopted. In this section, we will explore the Bayesian variant of the above policy gradient techniques which is based on modeling the policy gradient as a Gaussian Process which requires fewer samples to obtain an accurate gradient estimate. A Gaussian process (GP) is an indexed set of jointly Gaussian random variables. The Bayesian Policy Gradient technique considers the complete trajectories as the basic observable unit, hence it cannot take advantage of the Markov property (memoryless stochastic process) when the system is Markovian. In contrast, the BAC (Bayesian Actor Critic) method utilizes the Markov property and leverages Gaussian Process Temporal Differencing (GPTD). 


A GP prior is considered for action-value function, $Q(s,a)$, using a prior covariance kernel, then a GP posterior is computed sequentially on every single observed transition. GPTD is a learning model based on a generative model relating 
the observed reward to the unobserved action-value function given by:

\begin{equation}\label{eqn:GPTD}
r\left(z_{i}\right)=Q\left(z_{i}\right)-\gamma Q\left(z_{i+1}\right)+N\left(z_{i}, z_{i+1}\right)
\end{equation}

The Fisher kernel is considered as the prior covariance kernel for the GPTD state-action advantage function. For computing the gradient of the policy, the concept of Bayesian Quadrature is utilized, which we will explain in detail further.

The BAC algorithm is an extension of the Bayesian Policy Gradient (BPG) algorithm for learning policies of a continuous state space MDP problem, where the policy is updated by computing the gradient of the expected return given by:

\begin{equation}\label{exp_ret_grad}
\nabla_{\boldsymbol{\theta}} \eta(\boldsymbol{\theta})=\int \rho(\boldsymbol{\xi}) \frac{\nabla \operatorname{Pr}(\xi ; \boldsymbol{\theta})}{\operatorname{Pr}(\xi ; \boldsymbol{\theta})} \operatorname{Pr}(\boldsymbol{\xi} ; \boldsymbol{\theta}) d \boldsymbol{\xi}
\end{equation}

In the equation, the quantity $\frac{\nabla \operatorname{Pr}(\xi ; \boldsymbol{\theta})}{\operatorname{Pr}(\xi ; \boldsymbol{\theta})}$ is called the Fisher score function. Since the initial state probability distribution $P_{0}$ and the state transition probability $P$ are independent of the policy parameter $\boldsymbol{\theta}$ 
The example Gaussian policy distribution with parameter $\theta$ is given in the Section~\ref{bac_exp}. Evaluation of Bayesian Actor Critic Method, for a specific trajectory $\xi$, we can represent the Fisher score as:

\begin{equation}\label{grad_exp_return}
    u(\xi;\theta) = \frac{\nabla \operatorname{Pr}(\xi ; \boldsymbol{\theta})}{\operatorname{Pr}(\xi ; \boldsymbol{\theta})} = \sum_{t=0}^{T-1} \nabla \log \mu (a_{t} | x_{t}; \theta)
\end{equation}

The Monte Carlo method would utilize a number, say M, of sample paths $\xi_{i}$ and estimate the gradient $\nabla_{\boldsymbol{\theta}} \eta(\boldsymbol{\theta})$ using:

\begin{equation}
\widehat{\nabla \eta}(\theta)=\frac{1}{M} \sum_{i=1}^{M} R\left(\xi_{i}\right) \sum_{t=0}^{T_{i}-1} \nabla \log \mu\left(a_{t, i} \mid x_{t, i} ; \theta\right)
\end{equation}

where, $T_{i}$ is the $i^{th}$ episode length. The complete algorithm of the BAC is given by Alg.~\ref{bac_algo}, while Alg.~\ref{bac_grad} is the procedure to compute the gradient. The variables $k_x$ and $k_F$ introduced in the Alg.~\ref{bac_grad} are the state kernel and Fisher information kernel. The $\alpha$ and $C$ are the variables associated with the posterior Q computed in the BAC method (refer to~\cite{brl_survey} for the detailed description):

\begin{algorithm}[h]
	\caption{Pseudo Code for BAC}
	\begin{algorithmic}[1]
		\Function{$BAC$}{$N_u$,$N_e$}
		\State Initialize $\theta$
		 \For {$i$ in $N_u$}
		    \State Initialize Fisher Information Matrix, $G$
		    \If {$i\;\% \;u_{freq} \;== \;0$}
                    \State Evaluate policy $\theta$ for $e_{eval}$ episodes
            \EndIf
		     \For {$e$ in $N_e$}
		            \State $s$ = $env.reset()$
		            \State $u$ = $Fisher\_Score(s,theta)$ as per Eq.~\ref{grad_exp_return}
		            \State Till end of episode update $G$ = $G + u*u'$
		            \State store $s,u$ in $Episodes$
	            \EndFor
	            \State Compute $\Delta \theta = Gradient(Episodes,G)$
	            \State $\theta = \theta + \beta * \Delta \theta$
            \EndFor
		\EndFunction
	\end{algorithmic}
	\label{bac_algo}
\end{algorithm}

\begin{algorithm}[h]
	\caption{Pseudo Code for BAC Gradient}
	\begin{algorithmic}[1]
		\Function{$Gradient$}{$Episodes,G$}
		\For {$e$ in $Episodes$}
		    \State Compute $G^{-1}$
		    \State Fisher Info. Kernel $k_F = u'G^{-1}u$ \Comment{$u$ from $e$}
		    \State State kernel $k_x$ is computed using Gaussian func
		    \State $k = k_x +k_F$
		    \State Update $\alpha,C$ from the episode $e$ (Eq.~\ref{alpha_C})
	    \EndFor
	    \State Posterior mean and variance of $\delta \theta$ is computed to be $U*\alpha$ and $G - U*C*U'$, where $U$ is the Fisher score vector
		\EndFunction
	\end{algorithmic}
	\label{bac_grad}
\end{algorithm}

\section{Experimental Results}

In the AVC problem, the state space is represented by the bus voltages, while the action space is represented by the set-point of the generators. For the experimentation, two transmission electric grids: a) WSCC  and b) IEEE-14 bus case, with 3 and 5 generators respectively are considered. Among the two cases, WSCC case is considered for all the three BRL techniques, while IEEE-14 bus is evaluated for the BDQN technique.

\subsection{Evaluation of Bayesian Q Learning technique}

\textbf{\textit{Evaluate various exploration exploitation techniques for Bayesian Q Learning:}}
For the Bayesian Q learning technique, the experiments are performed to keep the voltage at the load buses at the permissible limit. In the WSCC case there are three load buses at Bus 5,6, and 8, among which the average load at Bus 6 is the largest with 185 MW, 
which is more challenging to regulate the voltage in comparison to bus 5 and 8 with the average of  150 and 100 MW respectively. Three different techniques of exploration-exploitation are considered:
\begin{enumerate}
    \item \textit{Q-Value Sampling}: This sampling technique is based on~\cite{wyatt_thesis}, where the actions are selected stochastically based on the current subjective belief of the action being optimal, i.e., the probability by which an action $a$ is performed is given by
    \begin{equation}
\operatorname{Pr}\left(a=\arg \max _{a^{\prime}} \mu_{s, a^{\prime}}\right)=\operatorname{Pr}\left(\forall a^{\prime} \neq a, \mu_{s, a}>\mu_{s, a^{\prime}}\right)
\end{equation}
\noindent where the $\mu_{s,a}$ is the mean estimate of the posterior distribution of $Q(s,a)$
    
    \item \textit{Greedy selection}: In this technique, the action $a$ that has the highest mean estimate for $Q(s,a)$ for state $s$ is selected.
    \item \textit{VPI}: It selects the action with highest expected return and VPI  i.e. $E[Q_{s,a}] + VPI(s,a)$ as described in Eq.~\ref{eqn:vpi}.
\end{enumerate}

\begin{figure}[h]
  \centering
  \subfigure[]{\includegraphics[height=1.5 in,width=1.5 in]{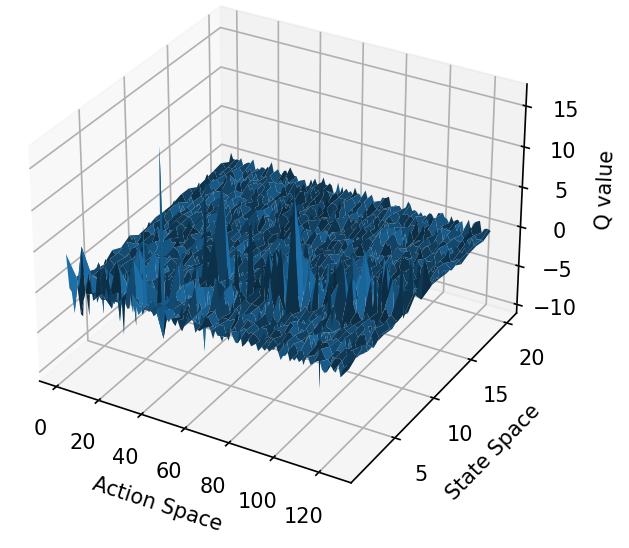}}
  \subfigure[]{\includegraphics[height=1.5 in,width=1.5 in]{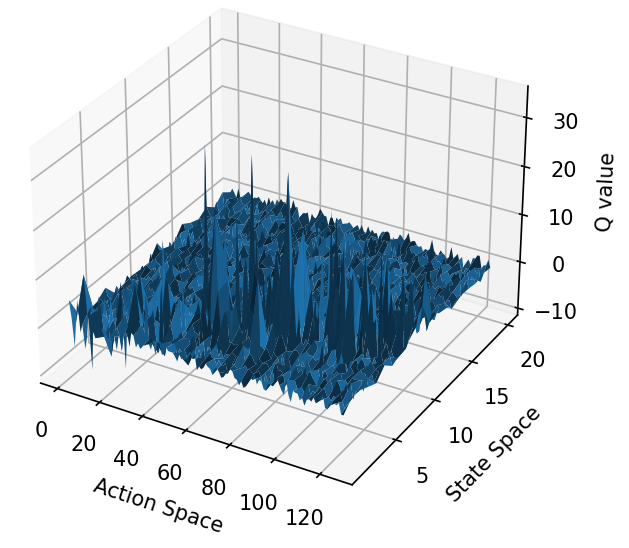}}
  \caption{Evolution of Q value distribution by opting for \textit{Q-Value Sampling} after (a) 1000 episodes  (b) 4000 episodes. 
  As more data samples are collected with more number of episodes, the Q values magnitude increases across the diagonal in the State-Action space plane. As the bus-voltage values goes down, the $AVC$ agent tries to reduce the set-point due to the ill-formed prior.
  }
  \label{fig:qval_dist}
  \vspace{-2mm}
\end{figure}

\begin{figure}[h]
  \centering
  \subfigure[]{\includegraphics[height=1.5 in,width=1.5 in]{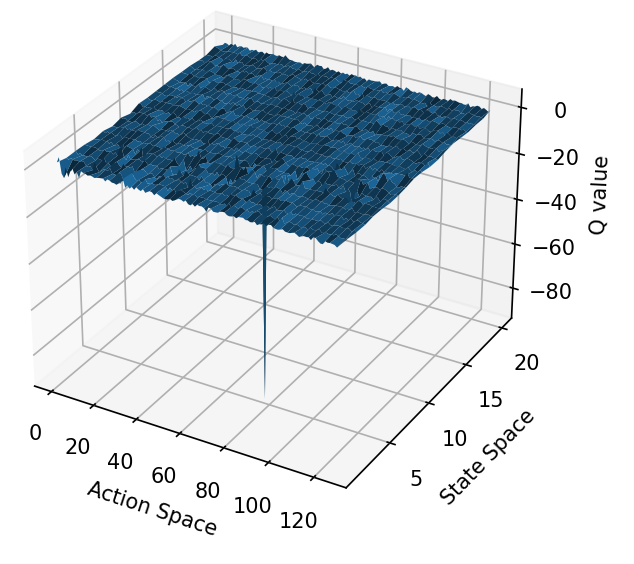}}
  \subfigure[]{\includegraphics[height=1.5 in,width=1.5 in]{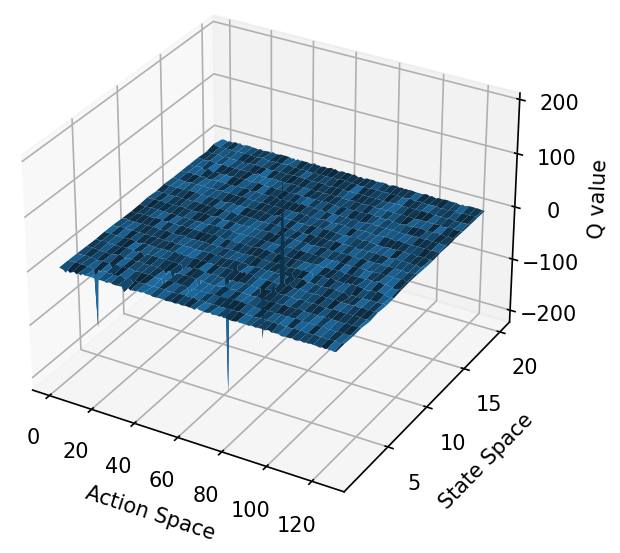}}
  \caption{
  Evolution of Q value distribution by opting for \textit{Greedy Selection} approach  after (a) 1000 episodes (b) 4000 episodes.  
  As more data samples are collected, with more number of episodes, the Q values magnitude in both positive and negative direction increases, hence may not be considered a good option for learning Q values.
  }
  \label{fig:greedy_dist}
  \vspace{-2mm}
\end{figure}

\begin{figure}[h]
  \centering
  \subfigure[]{\includegraphics[height=1.5 in,width=1.5 in]{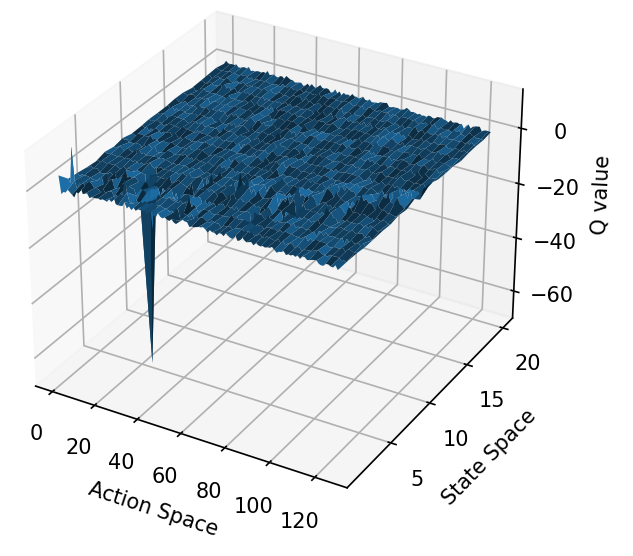}}
  \subfigure[]{\includegraphics[height=1.5 in,width=1.5 in]{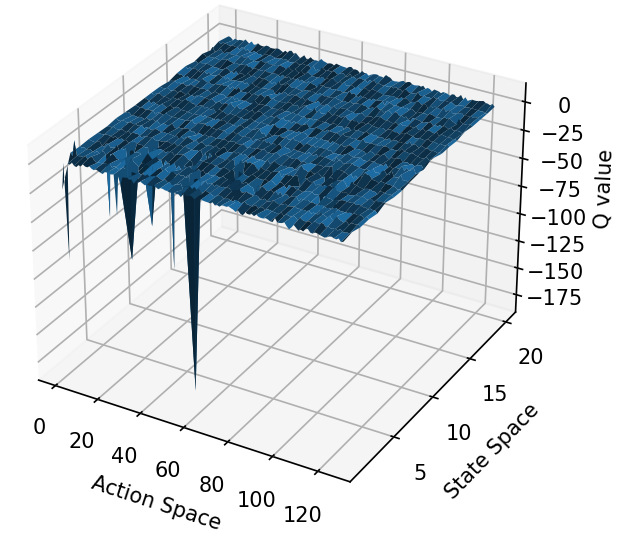}}
  \caption{
  Evolution of Q value distribution by opting for Q value + \textit{VPI} technique after (a) 1000 episodes (c) 4000 episodes.  
  The performance of the Q value + VPI technique is not as good in comparison to Q value sampling technique but it has lesser variance.
  }
  \label{fig:vpi_dist}
  \vspace{-2mm}
\end{figure}

\begin{figure*}[h]
  \centering
  \subfigure[]{\includegraphics[height=1.5 in,width=2.0 in]{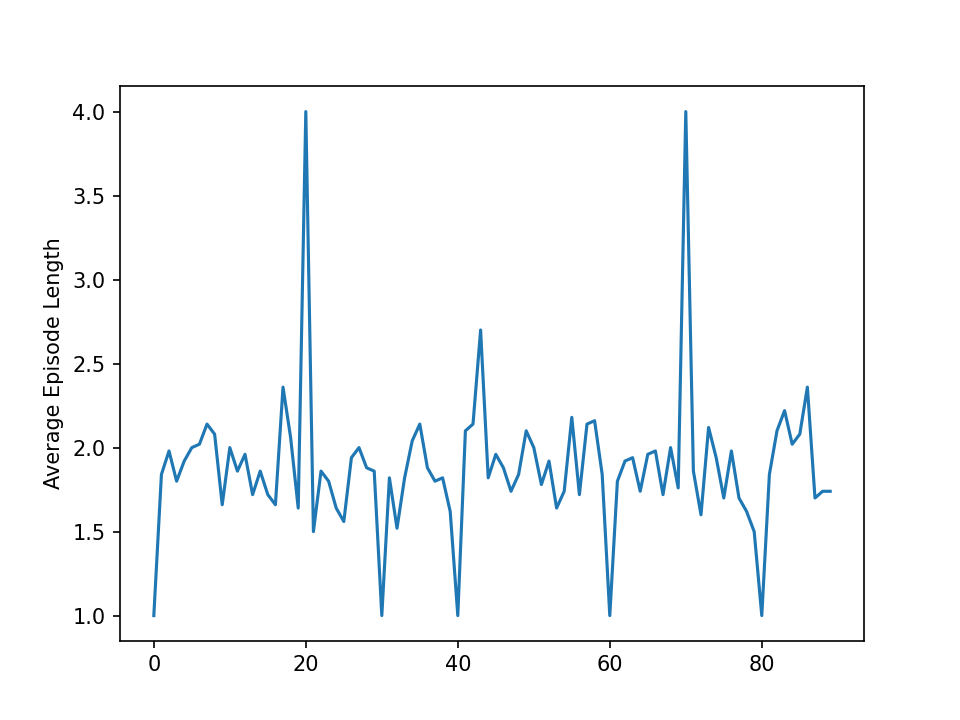}}
  \subfigure[]{\includegraphics[height=1.5 in,width=2.0 in]{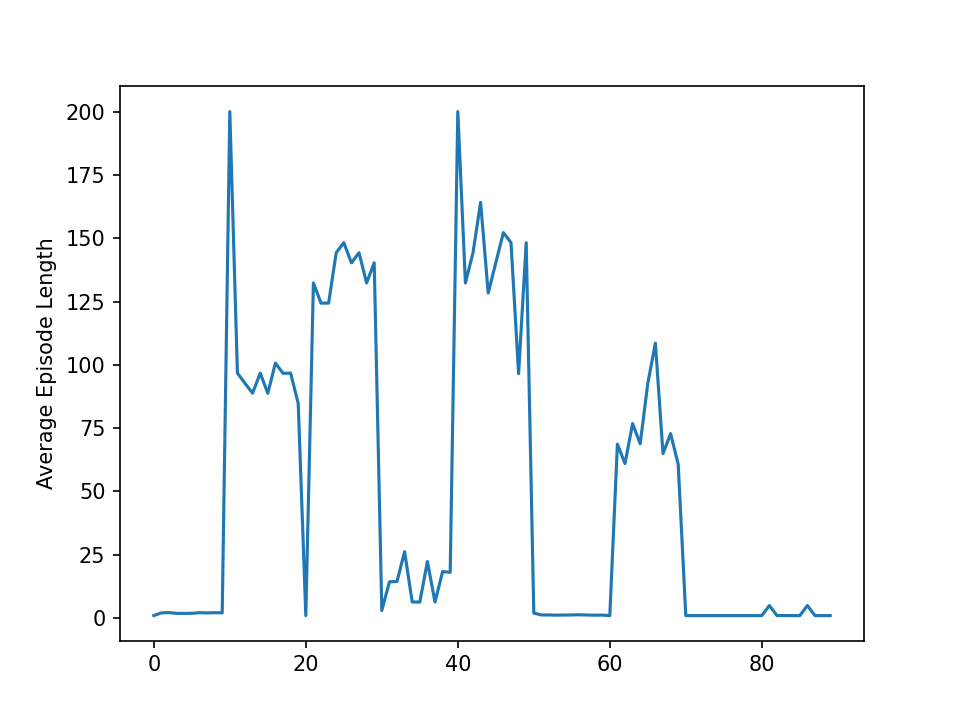}}
  \subfigure[]{\includegraphics[height=1.5 in,width=2.0 in]{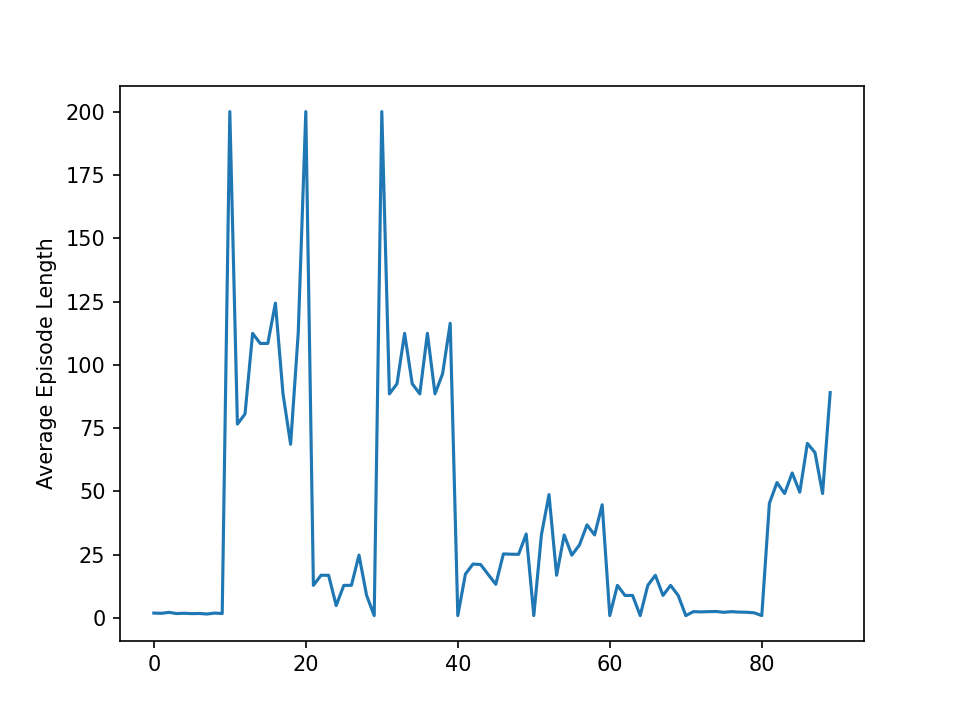}}
  \caption{Comparison of the average episode length for (a) Q value Sampling (b) Greedy Selection (c) Q Val + VPI. The X axis refers to every $50^{th}$ episode index and Y axis refer to the averaging over $50$ episode. 
  }
  \label{fig:compare_exp}
  \vspace{-2mm}
\end{figure*}

Fig.~\ref{fig:qval_dist}, Fig.~\ref{fig:greedy_dist}, and Fig.~\ref{fig:vpi_dist} show the distribution of the posterior Q value mean obtained with Q-value sampling, greedy, and VPI based techniques, respectively, obtained after completion of certain episodes. Fig.~\ref{fig:compare_exp} refers to the comparison of these techniques with respect to the average of the episode length to reach the goal state. Both \textit{Q-Value Sampling} and \textit{VPI} techniques perform better in terms of exploration, but Q-value sampling results in more variance. Based on the average episode length evaluation, the Q-value sampling technique does not result in reduction of episode length unlike the greedy selection and VPI techniques. Hence, based on the Q-value mean distribution as well as the average episode length, we preferred to use the VPI technique in further experiment on selection of better prior Q value distribution.

\begin{figure}[h]
  \centering
  \subfigure[]{\includegraphics[height=1.5 in,width=1.5 in]{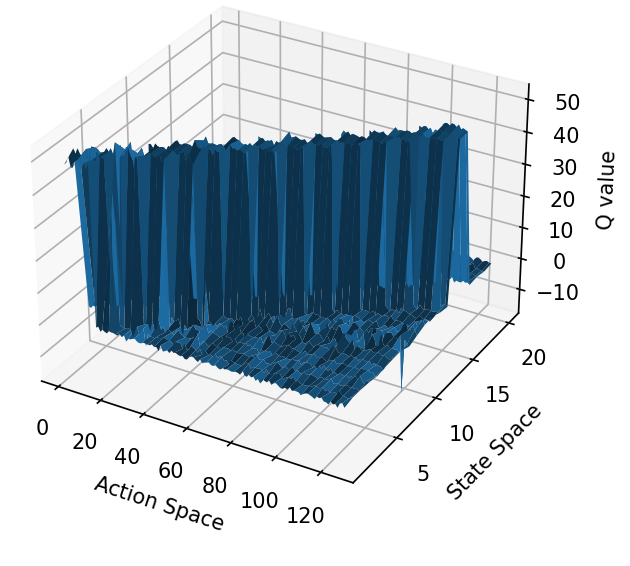}}
  \subfigure[]{\includegraphics[height=1.5 in,width=1.5 in]{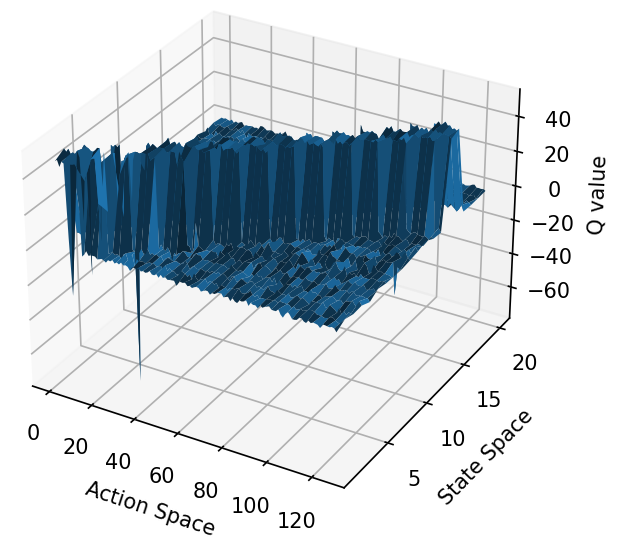}}
  \caption{
  Evolution of Q value distribution by opting for Q value + VPI technique but with an ill-formed prior after (a) 1000 episodes (b) 4000 episodes. The Q values magnitude are high across the diagonal in the State-Action space plane. As the bus-voltage values goes down, the $AVC$ agent reduces the set-point due to the ill-formed prior.
  }
  \label{fig:vpi_badprior}
\end{figure}

\begin{figure}[h]
  \centering
  \subfigure[]{\includegraphics[height=1.5 in,width=1.5 in]{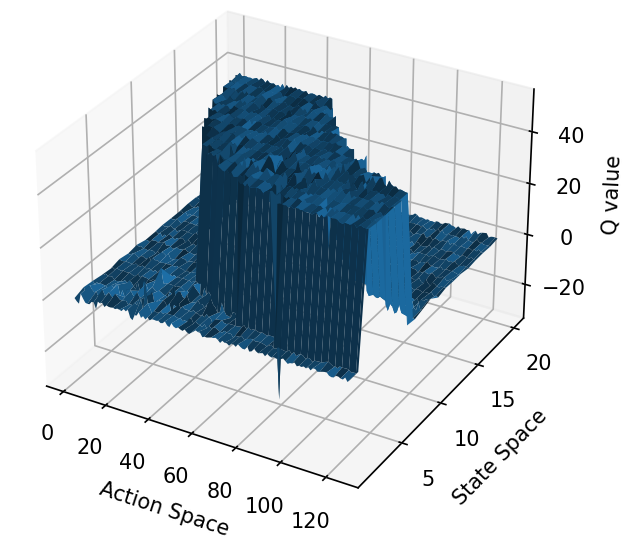}}
  \subfigure[]{\includegraphics[height=1.5 in,width=1.5 in]{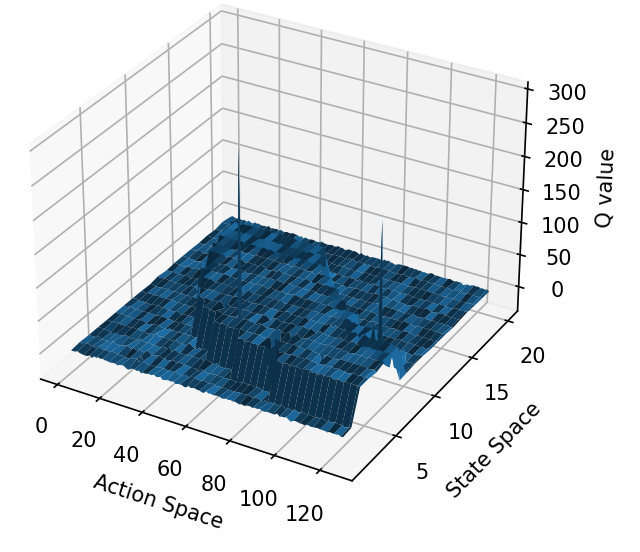}}
  \caption{
  Evolution of Q value distribution by opting for Q value + VPI technique but with an good prior after (a) 1000 episodes (b) 4000 episodes. The Q values magnitude are high across the cross-diagonal in the State-Action space plane. As the bus-voltage values goes down, the $AVC$ agent increases the set-point due to the good prior.
  }
  \label{fig:vpi_goodprior}
\end{figure}

\begin{figure*}[h]
  \centering
  \subfigure[]{\includegraphics[height=1.8 in,width=2.6 in]{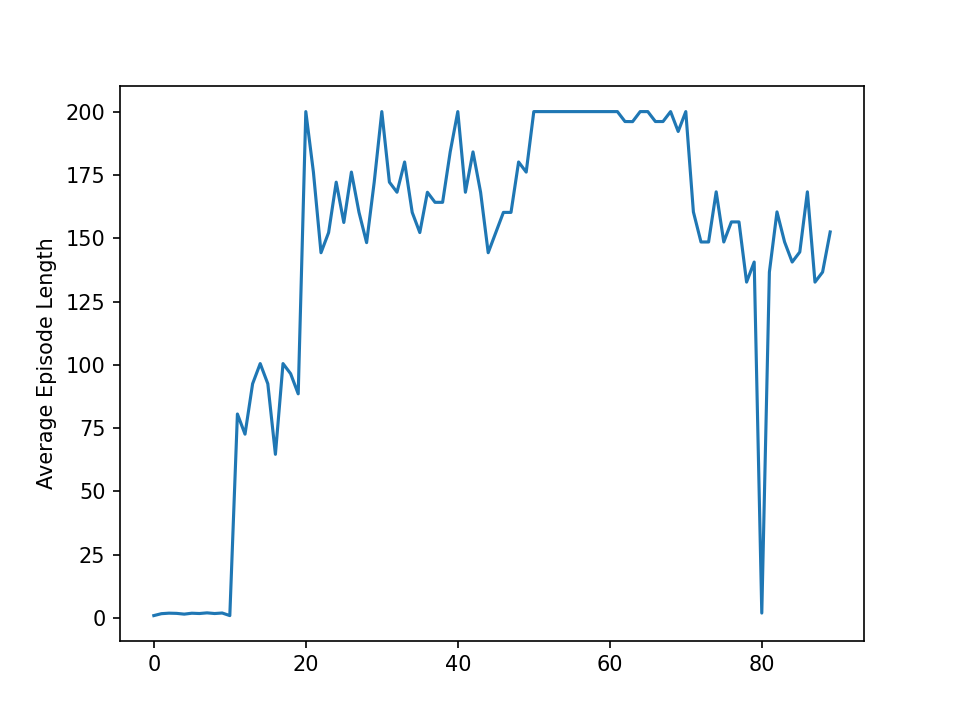}}
  \hspace{10mm}
  \subfigure[]{\includegraphics[height=1.8 in,width=2.6 in]{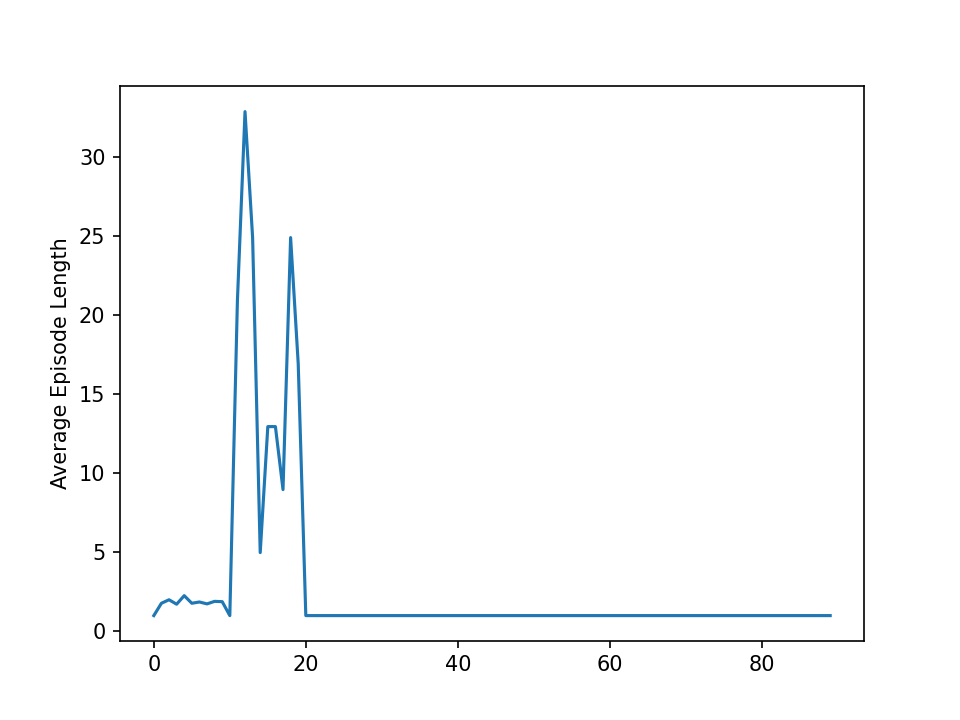}}
  \caption{Comparison of the average episode length for (a) Ill-formed Prior (b) Good Prior. The X axis refers to every $50^{th}$ episode index and Y axis refer to the averaging over $50$ episode. }
  \label{fig:compare_prior}
  \vspace{-7mm}
\end{figure*}

\begin{figure*}[h]
  \centering
  \subfigure[]{\includegraphics[height=1.8 in,width=2.6 in]{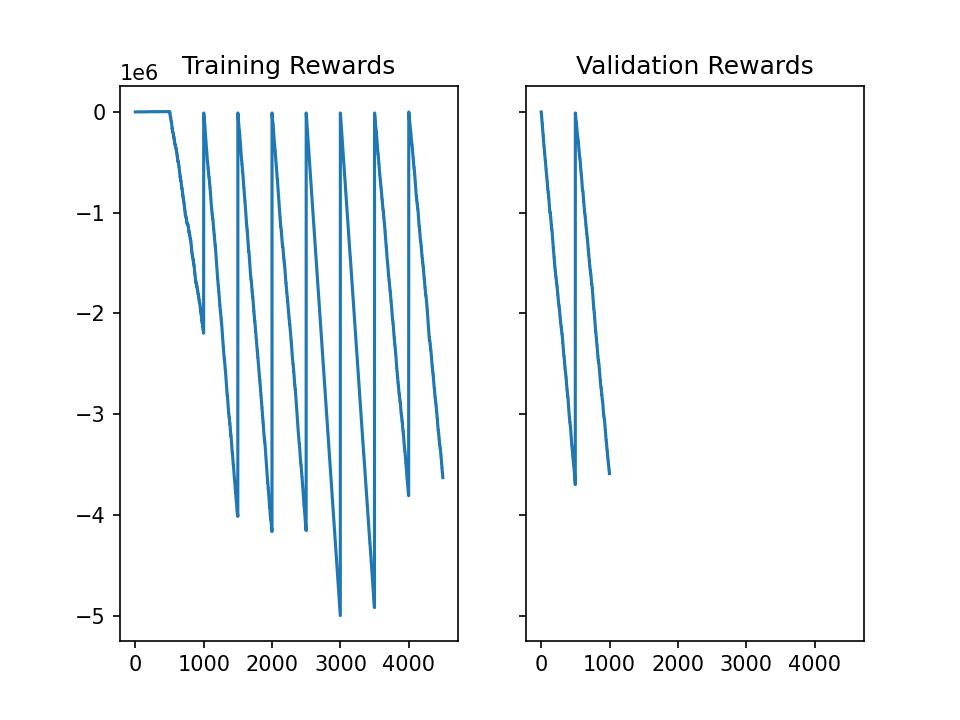}}
  \hspace{10mm}
  \subfigure[]{\includegraphics[height=1.8 in,width=2.6 in]{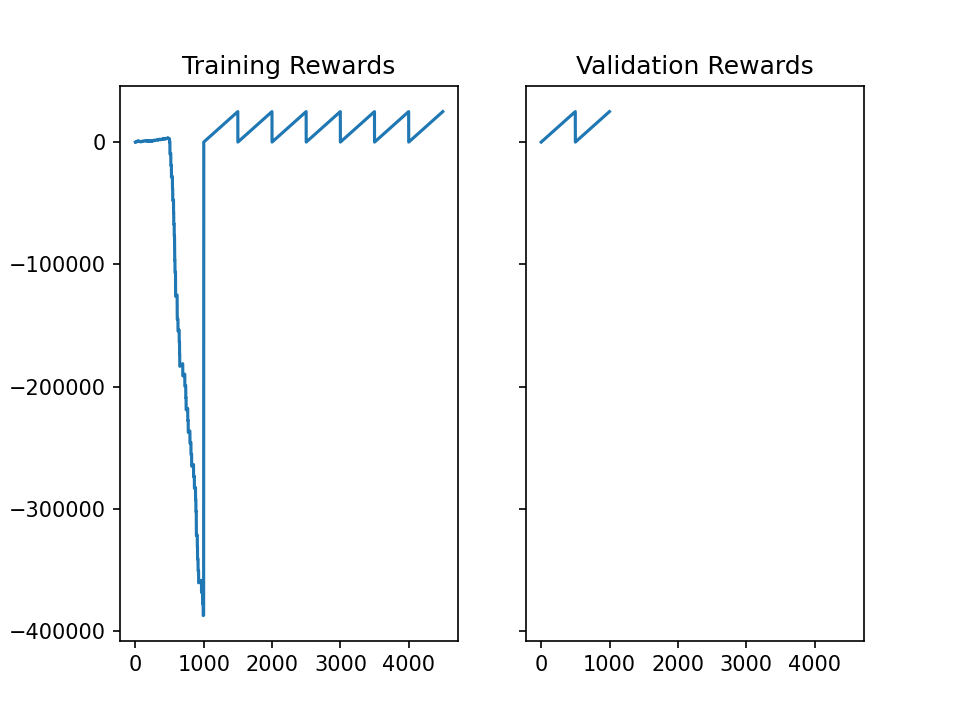}}
  \caption{Comparison of the average training and validation reward for (a) Ill-formed Prior (b) Good Prior. The X axis refers to the episode index. }
  \label{fig:compare_prior_reward}
\end{figure*}

\textbf{\textit{Evaluate the effect of selection of good prior for Bayesian Q Learning:}}
Further, we evaluate how a good prior $Q$ distribution can result in faster convergence in comparison to an ill-formed prior. The ill-formed prior refers to the function, where the actions would favor to lower the voltage set-point when the voltages are already low and to raise the voltage set-point when the voltages are already high; these are given higher Q values in the Q value prior distribution. By contrast, a ``good" prior is exactly opposite of the ill-formed prior. Fig.~\ref{fig:vpi_badprior} and Fig.~\ref{fig:vpi_goodprior} show the distribution of the posterior Q value mean using VPI method, after completion of certain episodes. 
While Fig.~\ref{fig:compare_prior} refers to the comparison of prior with respect to the average of the episode length to reach the goal state. Similar observations were made on the training and validation average rewards, as shown in Fig.~\ref{fig:compare_prior_reward}. Hence, selection of a better prior distribution can improve the performance as well as result in faster convergence in training the agent in very few episodes.

\subsection{Comparison of DQN with Bayesian DQN technique}
\begin{figure}[h]
\centering
  \includegraphics[width=0.8\columnwidth]{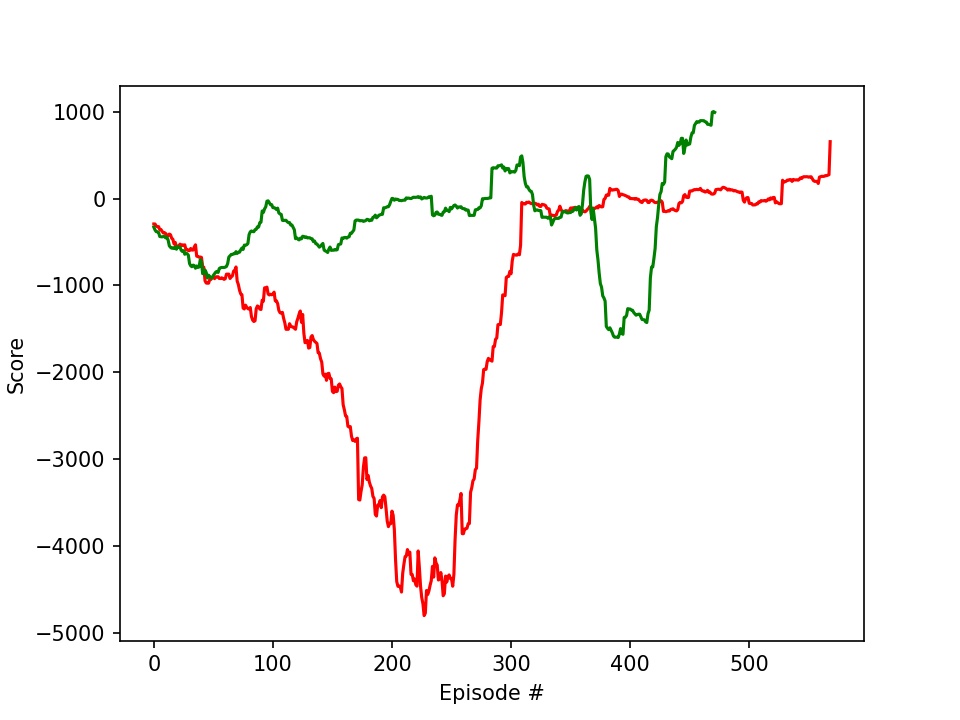}
  \caption{Comparison of BDQN \textit{(in green)} and DQN \textit{(in red)} agents with update frequency of 500 and 50000 samples used in MCMC based BDQN. The \textit{score} refer to the reward obtained in an episode while training. 
  }\label{fig:bdqn_vs_dqn}
\end{figure}

\begin{figure*}[h]
  \centering
  \subfigure[]{\includegraphics[width=0.40\linewidth]{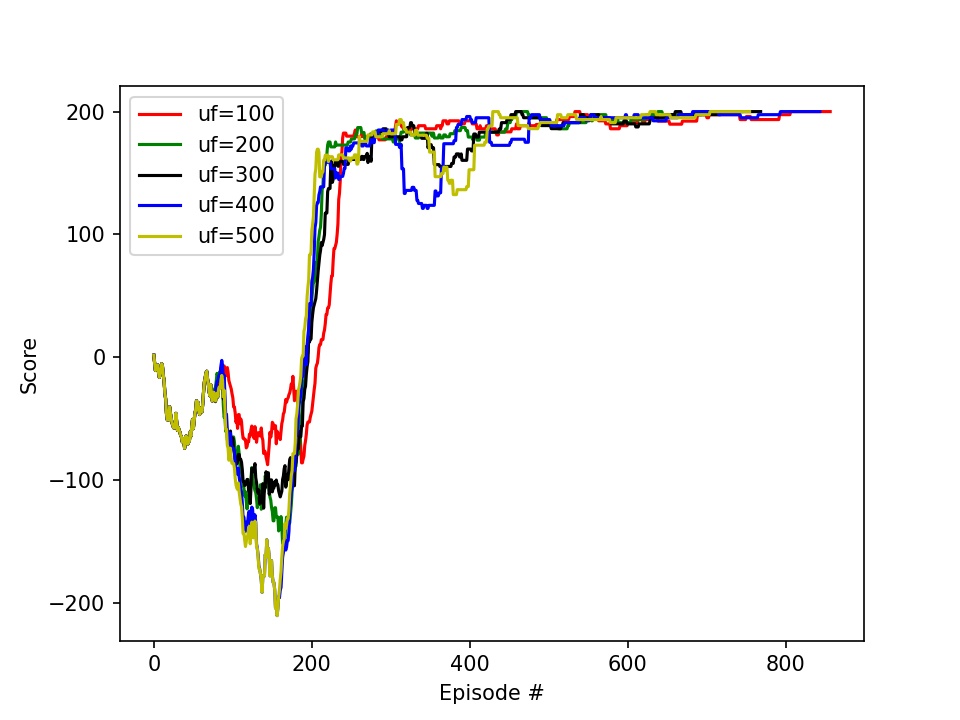}\label{dqn_varying_uf_ieee14}}
  \hspace{5mm}
  \subfigure[]{\includegraphics[width=0.40\linewidth]{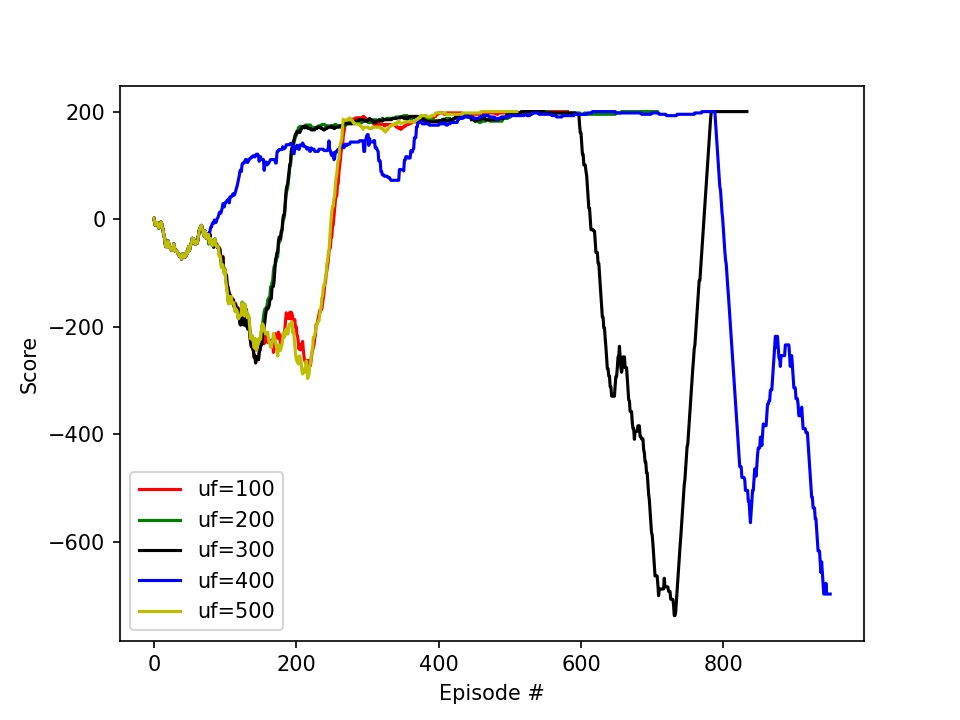}\label{bdqn_varying_uf_ieee14}}
  \caption{Evaluation of the effect of the policy update frequency ($uf$) for IEEE 14 bus systems using (a) DQN (b) BDQN. Lower the $uf$ value faster the target score of 200 is reached in BDQN. During the training, BDQN agent reaches the target goal score in less episodes in comparison to DQN.
  }
\end{figure*}

This section emphasizes on the results on how integration of BDQN agent (introduced in Section~\ref{bdqn_learn}) in comparison to DQN agent assist in training the agent utilizing less trajectories for voltage control. For the WSCC case, with 3 generators and 125 possible actions, Fig.~\ref{fig:bdqn_vs_dqn} shows how BDQN agent reached the desired goal of having an average reward of 200 in 420 steps while the DQN agent takes more than 580 steps, moreover the reward has higher variance in the DQN case. For the IEEE 14 bus case with 5 generators and 3125 possible actions too the BDQN reaches the goal of average reward of 200 in less number of episodes in comparison to the DQN (Figs.~\ref{dqn_varying_uf_ieee14} and~\ref{bdqn_varying_uf_ieee14}) for the update frequency $uf$ of 100, 200 and 500, except for the scenario of $uf=300,400$ in the BDQN case the average reward had higher variance. Unlike the WSCC case, in the IEEE 14 bus scenario, the network topology were altered in different scenarios, that is one of the probable reason for higher variance in the rewards.

\begin{figure}[h]
\centering
  \includegraphics[width=.9\columnwidth]{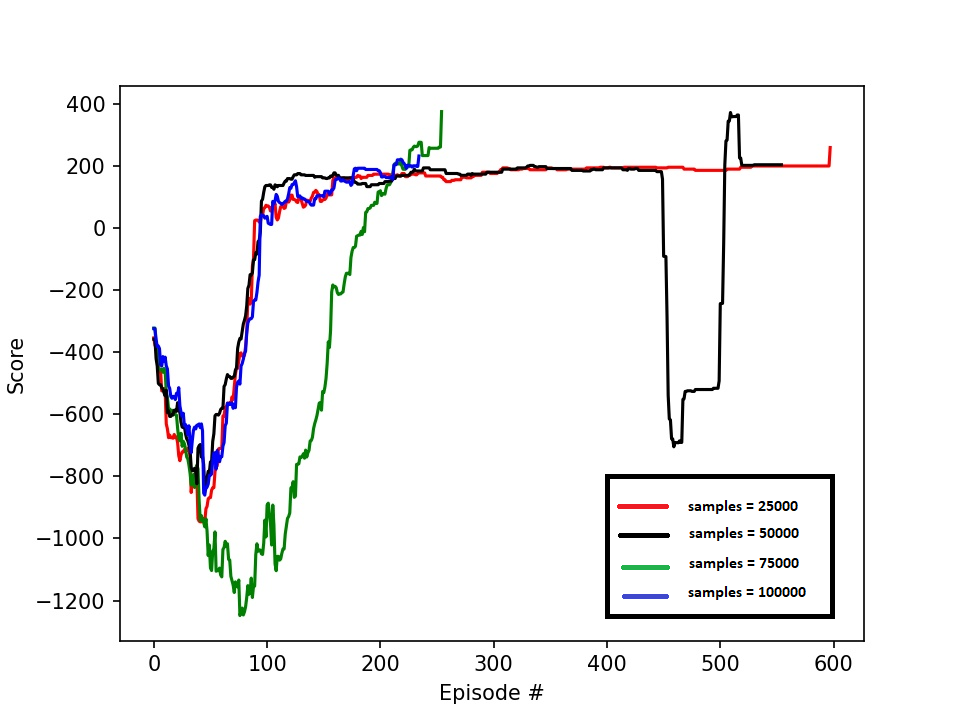}
  \caption{Evaluation of number of samples on the BDQN agent average reward with a fixed update frequency of 500 episodes. 
  }\label{fig:sample_effect_bdqn}
\end{figure}

We also evaluate the effect of changing the number of samples considered in the Metropolis Hasting Algorithms used in the MCMC method in BDQN learning. Fig.~\ref{fig:sample_effect_bdqn} shows how taking larger samples assist in reaching the goal in less number of training episodes. Similarly update frequency also effects the agent's training efficiency as shown in Table.~\ref{table:uf_bdqn}

\begin{table}[!h]
\begin{center}
\begin{tabular}{||cc||}
 \hline
Update Frequency & Episodes to reach goal  \\ \cline{1-2}
 100 &  111  \\ \cline{1-2}
 200 &  140  \\  \cline{1-2}
 300 &  172 \\  \cline{1-2}
 400 &  168 \\  \cline{1-2}
 500 &  197 \\  \cline{1-2}
\end{tabular}
\caption{Minimum number of episodes to train a BDQN agent with varying update frequency }
\end{center}
\label{table:uf_bdqn}
\end{table}


\subsection{Evaluation of Bayesian Actor Critic method}\label{bac_exp}
In this section, we evaluate the performance of the BAC method on the voltage control problem by taking a certain type of policy function and state space kernel. For the experiments, the policy used has the following form:
\begin{equation}
\mu\left(a_{i} \mid x\right)=\frac{\exp \left(\phi\left(x, a_{i}\right)^{\top} \theta\right)}{\sum_{j=1}^{125} \exp \left(\phi\left(x, a_{j}\right)^{\top} \boldsymbol{\theta}\right)}, \quad i=1,2,..,125
\end{equation}
where, 125 are the possible actions since there are 3 generators for the WSCC case with 5 discrete voltage levels of set-points. The policy feature vector 
$\phi(x,a_i) =(\phi(x)^{\top}\delta_{a_1,a_i},..., \phi(x)^{\top}\delta_{a_{125},a_i})$, 
where 
$\delta_{a_j,a_i}$ 
is 1, if 
$a_j = a_i$
and 0 otherwise.  While the state feature vector/ Gaussian State Kernel 
$\phi(x)$ 
is represented as follows:
\begin{equation}
\phi(x)=(\exp (\frac{-\|x-\bar{x}_{1}\|^{2}}{2 \sigma^2}), \ldots, \exp (\frac{-\|x-\bar{x}_{L}\|^{2}}{2 \sigma^2}))
\end{equation}
where, $L$ is the number of voltage levels the state is discretized and $\sigma^2$ refers to the variance of the Gaussian function. The $x$ is the p.u. voltage while the $x_1$ refers to the center of the first level in the voltage space and $x_L$ being the center of the last level. After every $u_{freq}$ (policy update frequency), the
learned policy is evaluated for $e_{eval} = 10$ episodes to estimate accurately the average number of steps to goal as well as the average episodic reward. We evaluate the performance by varying the voltage levels as well exploring the option with using different number of generators, to observe the effect of varying action space. The selection of variance $\sigma^2$ can also  effect the result, but currently this is not considered for evaluation. The BAC learning parameters are shown in Table~\ref{table:lp_bac}.

\begin{table}[h]
\begin{center}
\begin{tabular}{||c c c||}
\cline{1-3}
 Param & Desc & Val  \\ \cline{1-3}
  $N_{u}$ & Max. \# of policy updates &  200 \\ \cline{1-3}
 $u_{freq}$ & \# of policy updates per evaluation &  [5-25] \\\cline{1-3}
 $N_e$ & \# of episodes for every policy update  &  10  \\ \cline{1-3}
 $e_{eval}$ & \# of episodes in each evaluation &  10 \\ \cline{1-3}
 $\beta$ & Gradient Learning Rate & [.0025] \\  \cline{1-3}
 $e_{max}$ & Each episode maximum length &  10 \\  \cline{1-3}
 $\gamma$ & Discount Factor & 0.99 \\ \cline{1-3}
 $L$ & Voltage Discretization Level & [10,20] \\ \cline{1-3}
\end{tabular}
\caption{BAC Learning parameters}
\label{table:lp_bac}
\end{center}
\end{table}

\begin{figure*}[h]
  \centering
  \subfigure[]{\includegraphics[width=0.45\linewidth]{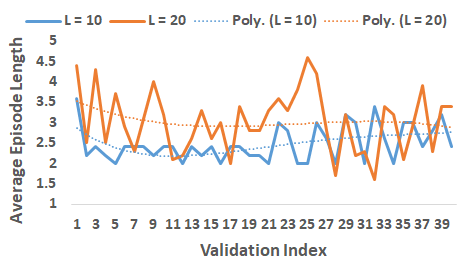}}
  \subfigure[]{\includegraphics[width=0.45\linewidth]{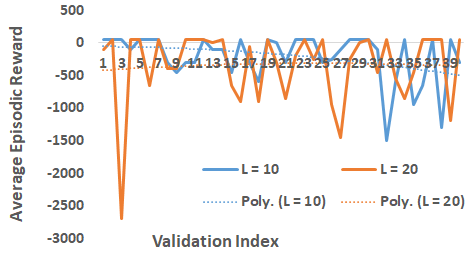}}
  \caption{Comparison of (a) Average Episode Length (b) Average Episodic Reward for two different voltage discretization levels $L=10$ and $L=20$ }
  \label{bac_volt_discretization}
\end{figure*}

\textit{\textbf{Evaluation of the voltage discretization level:}}
With increase in voltage levels, the state space increases so as the number of terms in the Gaussian State Kernel, which may cause more episodes for the policy to converge hence it can be observed from the Fig.~\ref{bac_volt_discretization}, the average number of episodes is less in the case of $L=10$ in comparison to $L=20$, but there is a decreasing trend in the case of $L=20$, which have better accuracy and less variance while proceeding towards obtaining optimal policy. Even the average episodic rewards have a more increasing trend with the scenario of $L=20$. Higher the levels the policy gradient technique can become computationally more expensive, but based on the improvement of performance $L=20$ is considered further to evaluate the effect of learning rate $\beta$ and update frequency $u_{freq}$ in the subsequent sub-section. The dotted lines in the figures, show the polynomial trendline.

\begin{figure*}[h]
  \centering
  \subfigure[]{\includegraphics[width=0.32\linewidth]{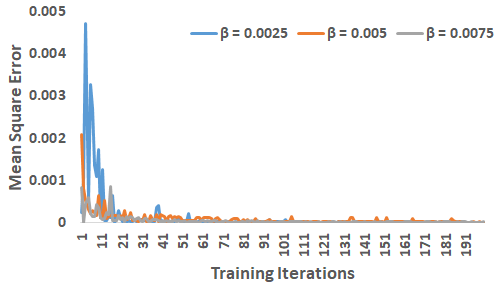}}
  \subfigure[]{\includegraphics[width=0.32\linewidth]{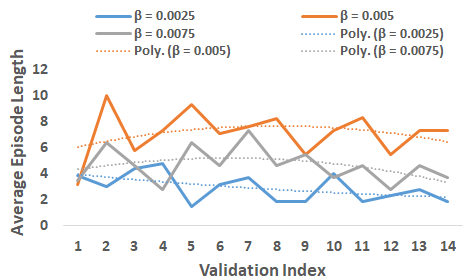}}
  \subfigure[]{\includegraphics[width=0.32\linewidth]{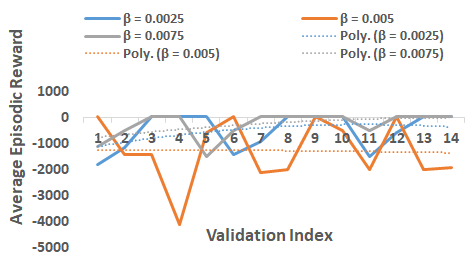}}
  \caption{Effect of varying learning rate, $\beta$, of the BAC policy gradient algorithm based of (a) Mean Square Error (b) Average Episode Length (c) Average Episodic Reward}
  \label{bac_lr_effect}
\end{figure*}

\textit{\textbf{Effect of BAC Learning Rate:}} The amount of change to the BAC model during each step, is the learning rate, $\beta$, and is one of the most important hyper-parameter for training. It determines how quickly or slowly the model learns. Higher learning rate generally allows the model to learn faster but results in sub-optimal policies. Fig.~\ref{bac_lr_effect} shows the effect of the learning rate $\beta$ on the mean square error (based on the deviation of the bus voltage values from 1.0 p.u.), average episode length and average episodic reward. Among three different $\beta$ values, $\beta=.0025$ is better and $\beta=.005$ is worse among all. Hence, we further consider $\beta=.0025$ for the experiments to study the effect of $u_{freq}$ of the BAC algorithm. The dotted lines in the figures, show the polynomial trendline.

\begin{figure*}[h]
  \centering
  \subfigure[]{\includegraphics[width=0.32\linewidth]{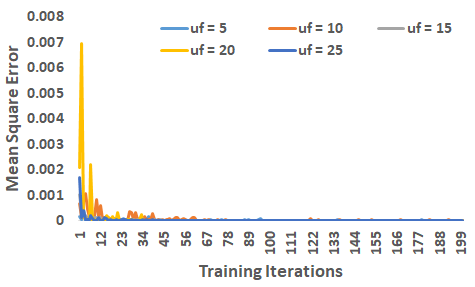}}
  \subfigure[]{\includegraphics[width=0.32\linewidth]{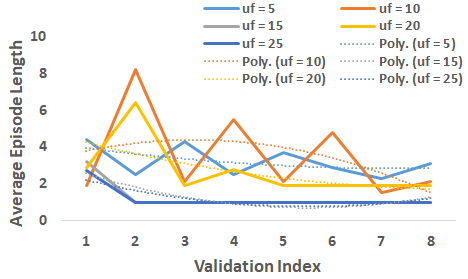}}
  \subfigure[]{\includegraphics[width=0.32\linewidth]{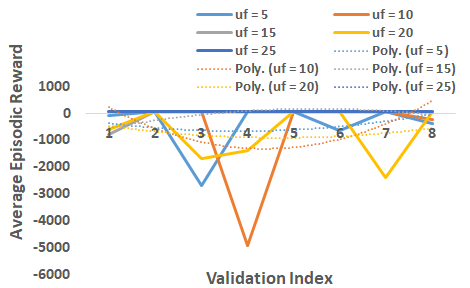}}
  \caption{Effect of varying update frequency $u_{freq}$ of the BAC policy gradient algorithm based of (a)Mean Square Error (b) Average Episode Length (c) Average Episodic Reward}
  \label{bac_uf_effect}
\end{figure*}

\textit{\textbf{Effect of BAC policy update frequency:}}
Line 5 in the Algorithm~\ref{bac_algo} shows the frequency with which we evaluate the trained policy parameter $\theta$. Fig.~\ref{bac_uf_effect} shows the effect of the update frequency, $u_{freq}$, on the mean square error, average episode length and average episodic reward. The $u_{freq}=15,25$ performs better with non-negative rewards and low average episode length, quite early in the validation index. Usually it is suggested to update with enough transitions captured through the episodes before doing a policy updates, but that would again depend on other factors such as how difficult the environment were. The results for the Bayesian Actor Critic (BAC) method can be further improved by considering other kernel functions and exploring more hyper-parameters. 

\section{Conclusion}
In this work, a POMDP for automatic voltage regulation in the transmission grid is formulated and solved using various BRL techniques. The major findings were: \textbf{a)} BDQN outperforms DQN agents in training the agents in less number of episodes. \textbf{b)} Value of Perfect Information based exploration-exploitation techniques improved the performance of the Bayesian-Q learning technique. \textbf{c)} Selection of an appropriate prior Q distribution plays a major role in the Bayesian Q learning process. \textbf{d)} Bayesian Actor Critic method is evaluated on the basis of the impact of hyper-parameters on the learning performance. 

\section*{Acknowledgement}
\label{Acknowledgement}
This work is supported by funds from the US Department of Energy under award DE-OE0000895.

\ifCLASSOPTIONcaptionsoff
  \newpage
\fi



%
\bibliographystyle{IEEEtran}
\bibliography{main}


%








\end{document}